\documentclass{article}

\usepackage[whole]{bxcjkjatype}
\usepackage{arxiv}

\usepackage[T1]{fontenc}
\usepackage[hidelinks]{hyperref}       
\usepackage{url}            
\usepackage{booktabs}       
\usepackage{amsfonts}       
\usepackage{nicefrac}       
\usepackage{microtype}      
\usepackage{lipsum}		
\usepackage{graphicx}
\usepackage{natbib}
\usepackage{doi}

\title{How does Burrows' Delta work on medieval Chinese poetic texts?}

\author{\href{https://orcid.org/0000-0002-9099-0436}{\includegraphics[scale=0.06]{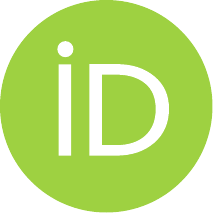}\hspace{1mm}Boris Orekhov}\footnote{webpage: \href{https://nevmenandr.github.io/homepage/}{https://nevmenandr.github.io/homepage/}, 	alternative email: nevmenandr@gmail.com} \\
	School of Linguistics\\
	HSE University,\\
	Institute of Russian Literature (Pushkin House)\\
	Russian Academy of Sciences \\
	\texttt{borekhov@hse.ru} 
}

\renewcommand{\shorttitle}{How does Burrows' Delta work on medieval Chinese poetic texts?}

\hypersetup{
pdftitle={How does Burrows' Delta work on medieval Chinese poetic texts?},
pdfsubject={cs.LG, cs.CL},
pdfauthor={Boris Orekhov},
pdfkeywords={tang poetry, burrows's delta, author attribution, chinese language},
}

\begin{document}
\maketitle

\begin{abstract}
	
	Burrows' Delta was introduced in 2002 and has proven to be an effective tool for author attribution. Despite the fact that these are different languages, they mostly belong to the same grammatical type and use the same graphic principle to convey speech in writing: a phonemic alphabet with word separation using spaces. The question I want to address in this article is how well this attribution method works with texts in a language with a different grammatical structure and a script based on different principles. There are fewer studies analyzing the effectiveness of the Delta method on Chinese texts than on texts in European languages. I believe that such a low level of attention to Delta from sinologists is due to the structure of the scientific field dedicated to medieval Chinese poetry. Clustering based on intertextual distances worked flawlessly. Delta produced results where clustering showed that the samples of one author were most similar to each other, and Delta never confused different poets. Despite the fact that I used an unconventional approach and applied the Delta method to a language poorly suited for it, the method demonstrated its effectiveness. Tang dynasty poets are correctly identified using Delta, and the empirical pattern observed for authors writing in European standard languages has been confirmed once again.
	
\end{abstract}


\section{Introduction}

Burrows' Delta was introduced in 2002 \citep{burrows2002delta} and has proven to be an effective tool for author attribution. The psychological and linguistic foundations of this method are not entirely clear, but numerous successful tests in various languages have confirmed that it is a viable method for calculating intertextual distance, which correlates well with a specific author's relationship to a text. If several texts are written by the same person, the Delta distance between them is smaller than between texts written by different authors.

Although this rule does not work in 100\% of cases \citep{skorinkin2023hacking}, it covers most cases that scholars in philology are interested in. The number of studies in which Burrows' Delta becomes a tool for solving attribution problems \href{https://scholar.google.ru/scholar?hl=ru&scisbd=1&as_sdt=0%2C5&q=Burrows%E2%80%99+Delta&btnG=}{is growing}. There are two reasons for this: first, the effectiveness of the method itself, repeatedly tested on texts in different languages. Second, the low entry threshold for researchers, provided by the well-crafted stylo package for the R language \citep{Eder2016Stylometry}, which generally follows a zero-coding strategy.

The effectiveness of the method has been proved on the material of different languages, both new and ancient: English \citep{hoover2004testing}, Old English \citep{garcia2006function}, German \citep{jannidis2014burrows}, Spanish \citep{hernandez2023challenging}, Italian \citep{savoy2018starnone, rybicki2018partners}, Polish \citep{rybicki2013stylistics}, Russian \citep{skorinkin2016specific}, Arabic \citep{abdulrazzaq2014burrows}, and others. There is a paper comparing the quality of attribution in different languages: Latin, Polish, English, German \citep{eder2011style}.

Despite the fact that these are different languages, they mostly (with the exception of Arabic) belong to the same grammatical type and use the same graphic principle to convey speech in writing: a phonemic alphabet with word separation using spaces. The question I want to address in this article is how well this attribution method works with texts in a language with a different grammatical structure and a script based on different principles.

There are fewer studies analyzing the effectiveness of the Delta method on Chinese texts than on texts in European languages. It is worth noting \citep{du2016testing} that those studies focus on modern Chinese, whereas I am interested in medieval Chinese poetry. This is a language without inflection and a script system without spaces as word delimiters. With such a grammatical structure, the distribution of function words is different from what is typical in European-standard languages \citep{Xiao2008OnTA, liu2017character}. Furthermore, in the writing systems familiar to Europeans, spaces separate words from each other. In Chinese, tokenization presents a complex problem \citep{huang2018pragmatic}, so it is important to check if Delta can be used by focusing only on the obvious text units, the characters, without resorting to complex algorithms that can introduce errors into text processing. Although the use of letter ngrams for Delta attribution is also found in European languages, typically sequences of several characters are used. Since in Chinese, the significance of a single character is higher than that of an individual letter in European languages, we will attempt to rely on individual characters rather than sequences.

I believe that such a low level of attention to Delta from sinologists is due to the structure of the scientific field dedicated to medieval Chinese poetry. Medieval Chinese literature is very different from medieval European literature, which had many gaps. The Chinese tradition of this period was highly regulated \citep{sturgeon2018unsupervised}, as was much of Chinese society, and it was built on the concept of authority, so texts were well documented. As a result, the material generally poses far fewer problems in terms of determining authorship. Authorship determination is the area where Delta traditionally shows the most significant results, which is why it is not in demand among sinologists specializing in medieval studies, as the authorship of texts is well known.

\section{Method and Data}
\label{sec:data}

There is no need to recount the calculation algorithm for computing Delta once again. This has been done dozens of times in the most well-known and authoritative publications. The intertextual distance is calculated using the formula:

\begin{equation}\label{eq:fourierrow}
	\Delta = \sum \limits_{i=1}^{n} \frac{|z(x_i) - z(y_i)|}{n} 
\end{equation}

Thus, we determine the distance between each and every text in the research sample. Since we do not have a reliable method for tokenizing the text, we are forced to work with individual characters. The Stylo package has an option that allows switching from words to letter n-grams. This means we treat each character as a separate letter, which is an intermediate approach between alphabetic scripts and the script characteristic of the Chinese language, where one word can be represented by multiple characters.

\subsection{Data}

As the source of texts, the collection stored in the repository at \citep{snowtraces2020} was used. It contains a digitized version of the "Complete Tang Poems" (or Quan Tangshi), which is the largest collection of Tang poetry. Thus, this study examines the effectiveness of the Delta method for determining authorship in Tang dynasty Chinese poetry.

I collected all texts of a single author together, resulting in 2537 poets. This is too many for a visual analysis of the results. Therefore, I further worked only with the 20 most prolific poets of the Tang era. Here is their list and the number of characters for each, including characters, spaces, line breaks, and punctuation marks:

\begin{enumerate}
  \item 白居易 229346
  \item 杜甫 129391
  \item 李白 103294
  \item 元稹 80968
  \item 韓愈 60013
  \item 劉禹錫 58224
  \item 貫休 48836
  \item 齊己 46381
  \item 陸龜蒙 45104
  \item 韋應物 40776
  \item 孟郊 40582
  \item 李商隱 39589
  \item 皎然 38349
  \item 劉長卿 35301
  \item 皮日休 34357
  \item 杜牧 33483
  \item 王建 31607
  \item 姚合 30887
  \item 錢起 29874
  \item 許渾 29775
\end{enumerate}

For stylometry, volume is important, so I combined the poems of one poet into several large samples to compare them with each other using Delta. However, the results can depend on which specific poems we combine. For example, one combination of poems might yield one version of stylometric distribution, while another combination might yield a different version. Therefore, it is necessary to try different combinations of poems within the sample. I did the following: I randomly mixed the order of poems for one author, split the volume in half, and presented each half as separate texts (samples), repeating this five times to create five test corpora. The code for these manipulations is provided in this repository \citep{Orekhov2024Delta}.

Next, I applied Delta to all these corpora using the Stylo package, configuring the package to work with characters instead of words and selecting the 100 most frequent characters. Here is the list of the 10 most frequent characters from the analysis:

\begin{enumerate}
  \item 不
  \item 人
  \item 無
  \item 一
  \item 日
  \item 山
  \item 風
  \item 有
  \item 何
  \item 來
\end{enumerate}

\section{Results}

\begin{figure}
	\centering
	\includegraphics[width=0.9\textwidth]{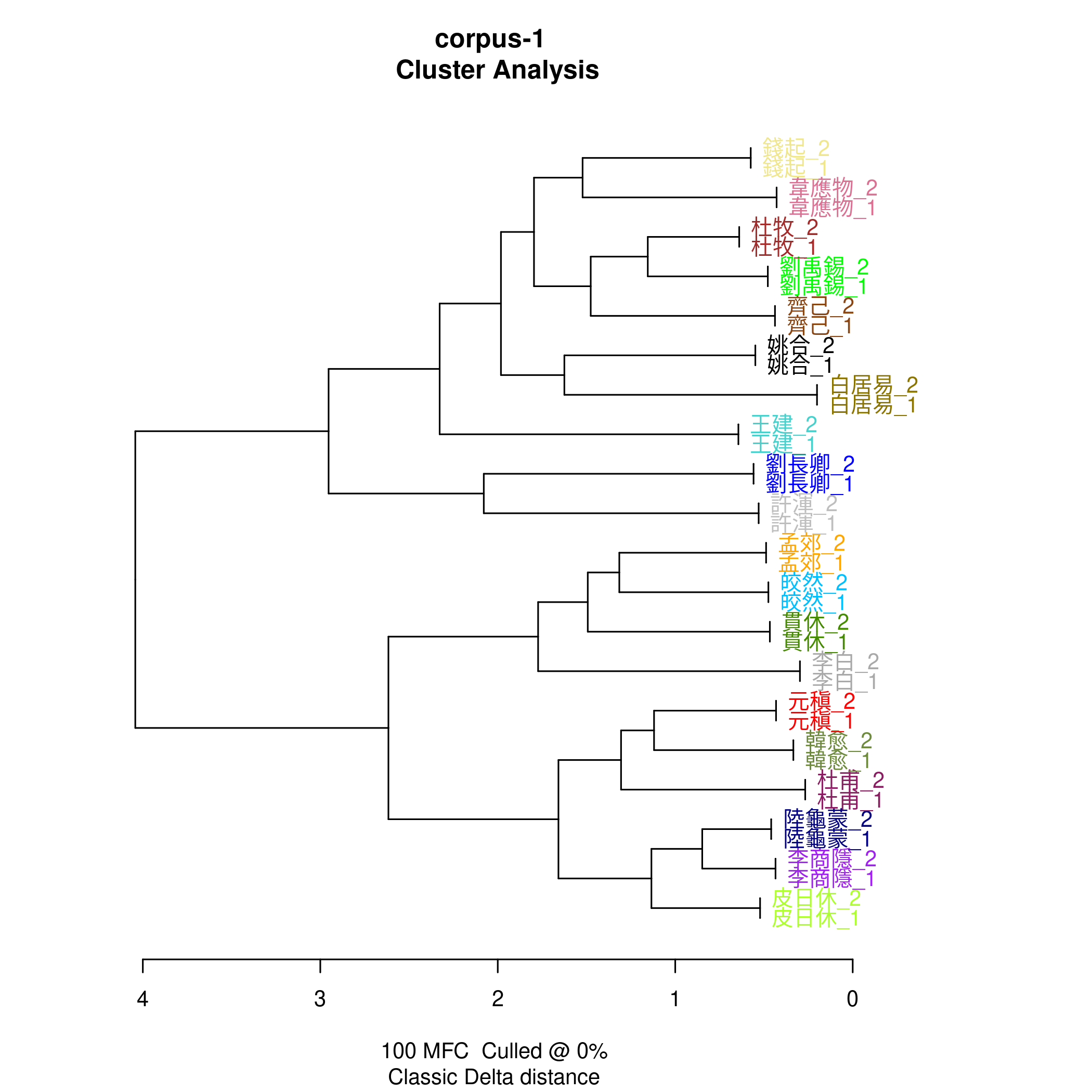}
	\caption{Cluster analysis of the first shuffled test corpus.}
	\label{fig:fig1}
\end{figure}

In all 5 versions of the corpus, clustering based on intertextual distances worked flawlessly. Delta produced results where clustering showed that the samples of one author were most similar to each other, and Delta never confused different poets. We present only one figure \ref{fig:fig1} as an illustration, as the others repeat the same result.

Let's provide a table \ref{tab:table} with a portion of the distances. The full table can be viewed in the data publication \citep{Orekhov2024Delta}.

\begin{table}[htbp]
    \centering
    \caption{Results: text distances}
    \begin{tabular}{llllll}
    \toprule
         & 元稹\_1 & 元稹\_2 & 劉禹錫\_1 & 劉禹錫\_2 & 劉長卿\_1\\ \midrule
        元稹\_1 & 0 & 0.4319 & 0.8917 & 0.8251 & 1.2953 \\ 
        元稹\_2 & 0.4319 & 0 & 0.9412 & 0.8734 & 1.3814 \\ 
        劉禹錫\_1 & 0.8917 & 0.9412 & 0 & 0.4782 & 1.1990 \\ 
        劉禹錫\_2 & 0.8251 & 0.8734 & 0.4782 & 0 & 1.2040 \\ 
        劉長卿\_1 & 1.2953 & 1.3814 & 1.1990 & 1.2040 & 0 \\ 
        劉長卿\_2 & 1.4421 & 1.4649 & 1.3361 & 1.3059 & 0.5586 \\ \bottomrule
    \end{tabular}
\label{tab:table}
\end{table}

\begin{figure}
	\centering
	\includegraphics[width=0.9\textwidth]{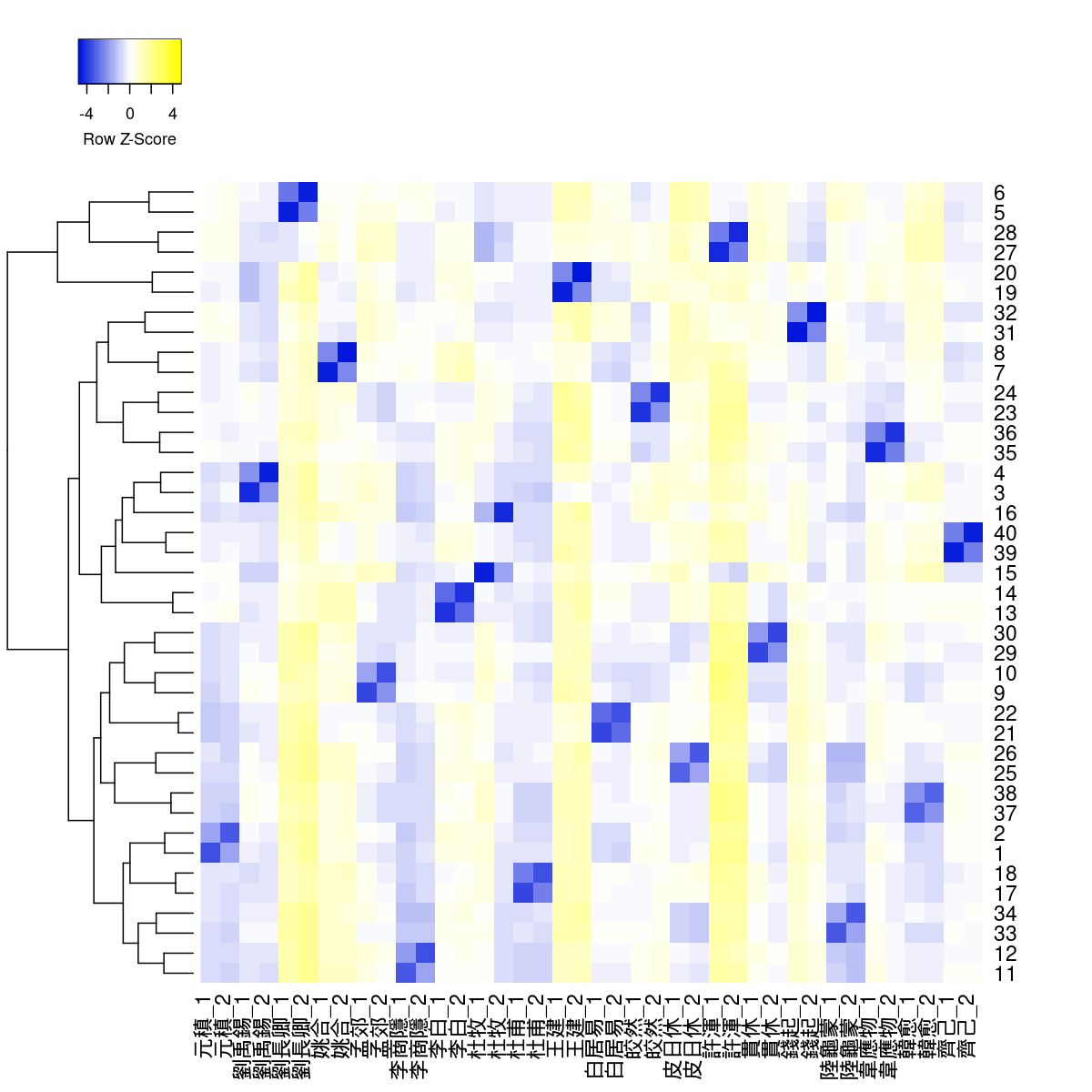}
	\caption{Heatmap of the delta text distances in the first shuffled test corpus.}
	\label{fig:fig2}
\end{figure}

The heatmap (Fig. \ref{fig:fig2}) of distances also records the smallest distances between samples of the same poet.

\section{Conclusion}

Despite the fact that I used an unconventional approach and applied the Delta method to a language poorly suited for it, the method demonstrated its effectiveness. Tang dynasty poets are correctly identified using Delta, and the empirical pattern observed for authors writing in European standard languages has been confirmed once again.

It should be noted that one character as a token is less than a word for medieval Chinese, but even 100 tokens were sufficient for accurate author clustering.

Further experiments can be conducted with a larger number of authors, different numbers of the most frequent tokens, and smaller sample sizes to understand the threshold at which the method ceases to work. However, the overall result is positive. It can be concluded that Delta works not only with European standard languages but also with such a regulated tradition, where the individuality of poetic style is tightly constrained by prescribed rules. Thus, an author's individual style can be hidden behind the rules imposed by tradition for expressing poetic ideas.

The confirmed effectiveness of the method has shown that raw texts can be used for Delta, as-is, without preprocessing with complex and error-prone tools like tokenizers.

If philologists need it, they can use Delta to solve attribution problems for medieval Chinese texts.

\section*{Acknowledgements}

I am grateful to Mariana Zorkina for her suggestions regarding the text corpus, specific bibliography points, and some observations about the research field, which we discussed in the early versions of this work. At the same time, all inaccuracies and errors in the text remain my responsibility.

\bibliographystyle{unsrtnat}
\bibliography{references} 

\end{document}